\documentclass[lettersize,journal]{IEEEtran}
\usepackage{amsmath,amsfonts}
\usepackage{algorithmic}
\usepackage{algorithm}
\usepackage{array}
\usepackage[caption=false,font=normalsize,labelfont=sf,textfont=sf]{subfig}
\usepackage{textcomp}
\usepackage{stfloats}
\usepackage{url}
\usepackage{verbatim}
\usepackage{graphicx}
\usepackage{cite}
\usepackage[numbers]{natbib}
\usepackage{tabularx}
\usepackage{listings} 
\usepackage{booktabs}

\usepackage{pifont}
\usepackage{graphicx}
\usepackage{enumitem}

\begin{document}

\title{Integrating Graphs, Large Language Models, and Agents: Reasoning and Retrieval}

\author{
Hamed Jelodar, 
Samita Bai,
Mohammad Meymani,
Parisa Hamedi,
Roozbeh Razavi-Far,
Ali Ghorbani
\thanks{Hamed Jelodar, Samita Bai, Mohammad Meymani, Parisa Hamedi, Roozbeh Razavi-Far, and Ali Ghorbani are with the Canadian Institute for Cybersecurity, Faculty of Computer Science, University of New Brunswick, Fredericton, NB E3B 9W4, Canada (e-mails: h.jelodar@unb.ca; samita.bai@unb.ca;mohammad.meymani79@unb.ca parisa.hamedi@unb.ca; roozbeh.razavi-far@unb.ca; ghorbani@unb.ca).}
\thanks{Hamed Jelodar contributed to conceptualization, methodology, and writing. Samita Bai, Parisa Hamedi, and Mohammad Meymani contributed to writing and editing. Roozbeh Razavi-Far and Ali Ghorbani contributed to project management, conceptualization, and writing.}
}



\markboth{Journal of \LaTeX\ Class Files,~Vol.~14, No.~8, August~2021}%
{Shell \MakeLowercase{\textit{et al.}}: A Sample Article Using IEEEtran.cls for IEEE Journals}


\maketitle

\begin{abstract}
Generative AI, particularly Large Language Models, increasingly integrates graph-based representations to enhance reasoning, retrieval, and structured decision-making. Despite rapid advances, there remains limited clarity regarding when, why, where, and what types of graph–LLM integrations are most appropriate across applications. This survey provides a concise, structured overview of the design choices underlying the integration of graphs with LLMs. We categorize existing methods based on their purpose (e.g., reasoning, retrieval, generation, recommendation), graph modality (knowledge graphs, scene graphs, interaction graphs, causal graphs, dependency graphs), and integration strategies (prompting, augmentation, training, or agent-based use). By mapping representative works across domains such as cybersecurity, healthcare, materials science, finance, robotics, and multimodal environments, we highlight the strengths, limitations, and best-fit scenarios for each technique. This survey aims to offer researchers a practical guide for selecting the most suitable graph–LLM approach depending on task requirements, data characteristics, and reasoning complexity.
\end{abstract}


\section{Introduction}
\IEEEPARstart{G}{enerative} Artificial Intelligence (Gen-AI) has rapidly transformed the landscape of intelligent systems, with large language models (LLMs) demonstrating remarkable capabilities in understanding, generating, and reasoning over unstructured text. Despite these advances, LLMs primarily operate on sequential token representations and often lack explicit mechanisms for modeling structured relational knowledge \cite{yang2026graph, he2026large, zhu2026knowpath}. In parallel, graph representations provide a powerful framework for capturing entities, relationships, and topological dependencies in structured data \cite{yang2026cti}. Consequently, integrating LLMs with graph-based learning has emerged as a promising direction to enhance reasoning, retrieval, and decision-making.\par
Graph-enhanced LLMs combine the semantic reasoning power of language models with the structured inductive bias of graphs. While LLMs excel at extracting contextual meaning from text, they often struggle to explicitly model complex relationships and multi-hop dependencies \cite{zhang2026scenellm}. Graph structures address this limitation by encoding entities and their interactions in an organized and interpretable manner. Based on previous works \cite{yu2026graphpilot, xiao2026reliable, tong2026gnn}, graph-LLM integration enables more reliable reasoning over structured knowledge that may be difficult to infer from text alone.\par
\begin{figure*}
    \centering
    \includegraphics[width=\textwidth]{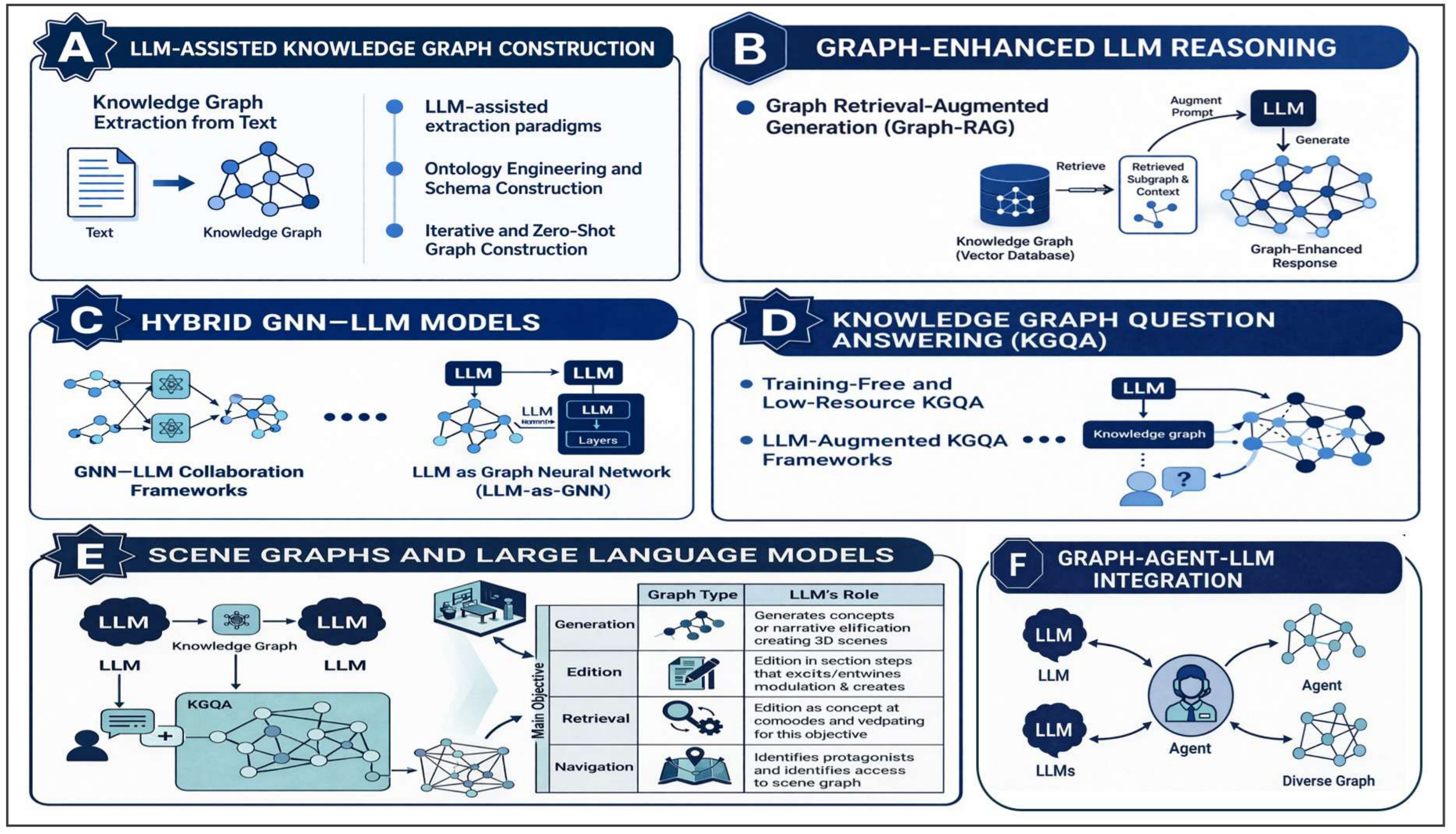}
    \caption{Overview of recent methods for integrating graphs and large language models (LLMs) across key paradigms}
    \label{fig1:integration}
\end{figure*}

Recent studies further demonstrate the effectiveness of this combination across multiple domains. For instance, applications span software engineering \cite{tao2025code}, recommendation systems \cite{ahn2026enriching}, healthcare \cite{li2025approximate} and clinical decision support \cite{park2025synergistic}, traffic prediction \cite{liu2025st}, and cybersecurity \cite{erlemann2026full, faye2026tegra}. For example, \cite{xu2026automated} introduces a graph-augmented LLM framework that incorporates relational structures to improve reasoning and predictive performance on structured datasets. Similarly, \cite{chen2025combining} proposes a hybrid LLM–GNN architecture in which semantic features generated by LLMs enhance graph-based classification and inference, highlighting the synergy between language and graph representations. Fig.~\ref{fig1:integration} illustrates the recent methods for integrating graphs and LLMs.

Motivated by these advances, this paper investigates recent strategies for integrating large language models with graph-based methods. Furthermore, we provide a structured overview of existing approaches and present a unified perspective on their design choices, capabilities, and application scenarios.\par
\noindent\textbf{Motivation}
Since 2018, the importance of transformer-based models and large language models (LLMs) has increased rapidly and continues to be highlighted across diverse real-world applications. More recently, integrating LLMs with graph-based representations has shown strong potential to produce more informative and structured reasoning outcomes. Although a limited number of studies have explored the use of LLMs in conjunction with graphs, we have not identified any systematic research that comprehensively investigates the different aspects of graph-LLM integration. In particular, there is a lack of studies addressing the fundamental questions (when, why, where, and what), regarding the effective use of LLMs with graph structures. In this research, we aim to provide a structured and systematic analysis of graph-LLM integration, clarifying when such integration is most beneficial and why graph representations enhance LLM reasoning.
We further examine where these approaches are most effective across application domains, and identify the key design choices and architectural patterns critical for their successful deployment.

\noindent\textbf{Why Combine Graphs and LLMs?}
To address the question of why graphs and large language models should be combined, the rationale largely depends on the specific use case and application requirements. LLMs are highly effective at understanding unstructured text, while graph-based models are effective at capturing structured relationships and dependencies.\\
For example, in software engineering \cite{tao2025code}, LLMs can analyze source code to infer semantic intent, whereas graphs can represent program structures such as call graphs and control-flow graphs. Together, they enable more accurate tasks such as code localization, vulnerability analysis, and program understanding.\\
Similarly, in healthcare and clinical decision support systems \cite{li2025approximate}, LLMs can interpret clinical notes, while graphs encode structured medical knowledge, including disease-symptom relationships and drug interactions. Their integration supports more reliable clinical reasoning, improved decision support, and reduces the risk of hallucination in high-stakes medical applications.

\noindent\textbf{Survey Scope and Contributions}
This survey focuses on the systematic analysis of recent efforts that integrate large language models with graph-based representations. We cover a broad range of graph modalities, including knowledge graphs, program and dependency graphs, interaction graphs, causal graphs, and multi-modal graph structures, and examine how they are combined with LLMs across diverse tasks and domains. The main contributions of this survey are summarized as follows:\\
\begin{itemize}
    \item We present a unified taxonomy of graph-LLM integration strategies, categorizing existing approaches by their functional role, graph modality, and integration mechanism.
    \item We investigate representative methods across multiple application domains, highlighting their strengths, limitations, and suitability for different task settings.
    \item We identify common design patterns and recurring challenges, including scalability, hallucination mitigation, interpretability, and reasoning depth.
    \item We provide practical guidelines to help researchers and practitioners select an appropriate graph-LLM integration approach based on the task complexity, data availability, and application constraints.
\end{itemize}
The rest of the paper is organized as follows. Section 2 presents the foundational concepts underlying Large Language Models (LLMs) and Graph Neural Networks (GNNs), highlighting their complementary strengths and limitations. Section 3 reviews LLM-assisted graph construction techniques, including knowledge graph extraction, ontology engineering, and iterative graph building methods. Section 4 introduces graph-enhanced LLM reasoning paradigms, with a particular focus on Graph Retrieval-Augmented Generation (GraphRAG) and related reasoning frameworks. Section 5 discusses hybrid GNN–LLM models, categorizing them into collaboration frameworks, directional integrations, explainability-enhanced methods, and pre-trained architectures. \\Section 6 explores knowledge graph question answering (KGQA) approaches, covering both training-free and LLM-augmented methods. Section 7 examines the integration of scene graphs with LLMs, including applications in generation, editing, retrieval, and navigation tasks. Section 8 presents graph–agent–LLM integration frameworks, emphasizing agentic reasoning and multi-step workflows. Section 9 highlights real-world applications across domains such as cybersecurity, healthcare, recommendation systems, and governance. Finally, the paper concludes with a summary of key insights, challenges, and future research directions.

\section{Foundations}
This section provides the background necessary to understand the integration paradigms discussed in the remainder of the paper. In particular, we review the two main technical foundations of this survey: large language models, which excel at semantic understanding and generation over unstructured text, and graph neural networks, which are designed to model structured entities and relationships. This contrast helps clarify why combining graphs with LLMs can improve reasoning, retrieval, and decision-making in complex tasks.

\subsection{Large Language Models}
Large Language Models are deep neural architectures trained on large-scale textual corpora to learn general-purpose representations of language. They capture statistical patterns, semantic relationships, and contextual dependencies across tokens, enabling a wide range of tasks including text generation, question answering, and reasoning. Most modern LLMs are built upon the Transformer architecture, which leverages self-attention mechanisms to model long-range dependencies without relying on recurrence. Due to large-scale pretraining, LLMs exhibit strong generalization and in-context learning capabilities \cite{moia2026llm, li2025generation}, allowing them to adapt to new tasks with minimal or no parameter updates.

Formally, let a tokenized sequence be denoted as 
\[
\mathbf{x} = (x_1, x_2, \dots, x_T),
\]
where each $x_t \in \mathcal{V}$ and $\mathcal{V}$ is the vocabulary. An LLM parameterized by $\theta$ models the joint probability of the sequence using the autoregressive factorization:
\[
P_\theta(\mathbf{x}) = \prod_{t=1}^{T} P_\theta(x_t \mid x_{<t}),
\]
where $x_{<t} = (x_1, \dots, x_{t-1})$. The model is trained by maximizing the likelihood over a dataset $\mathcal{D}$:
\[
\theta^* = \arg\max_{\theta} \; \mathbb{E}_{\mathbf{x} \sim \mathcal{D}} \left[ \sum_{t=1}^{T} \log P_\theta(x_t \mid x_{<t}) \right].
\]

The core computation in Transformer-based LLMs is the self-attention mechanism. Given an input representation $X \in \mathbb{R}^{T \times d}$, attention is computed as:
\[
\text{Attention}(Q, K, V) = \text{softmax}\left(\frac{QK^\top}{\sqrt{d_k}}\right)V,
\]
where 
\[
Q = XW_Q, \quad K = XW_K, \quad V = XW_V,
\]
and $W_Q, W_K, W_V$ are learnable projection matrices. Multiple attention heads are combined to capture diverse contextual interactions.

At inference time, given a context or prompt $\mathcal{C}$, the model generates an output sequence $\mathbf{y}$ as:
\[
P_\theta(\mathbf{y} \mid \mathcal{C}) = \prod_{t=1}^{|\mathbf{y}|} P_\theta(y_t \mid \mathcal{C}, y_{<t}),
\]
enabling in-context learning without modifying model parameters.

Despite their strong representational power, LLMs encode knowledge implicitly within parameters $\theta$, lacking explicit structured representations such as graphs $\mathcal{G} = (\mathcal{V}, \mathcal{E})$. This limits their effectiveness in tasks requiring explicit relational reasoning, multi-hop inference, and verifiable factual consistency.

\subsection{Graph Neural Networks}
Graph Neural Networks (GNNs) are neural architectures designed to operate on graph-structured data, where entities are represented as nodes and relationships are modeled as edges. Unlike sequential models, GNNs explicitly capture relational dependencies and structural patterns by propagating information over the graph topology. This makes them particularly effective for tasks involving structured reasoning, such as social network analysis, recommendation systems, biological networks, and program analysis.

The core mechanism of GNNs is \textit{message passing}, in which each node iteratively aggregates information from its local neighborhood to update its representation. Formally, let $\mathcal{G} = (\mathcal{V}, \mathcal{E})$ denote a graph with node set $\mathcal{V}$ and edge set $\mathcal{E}$. The representation of a node $v \in \mathcal{V}$ at layer $l$ is denoted by $\mathbf{h}_v^{(l)}$. A general message-passing layer can be written as:

\begin{equation}
\mathbf{h}_v^{(l+1)} = \phi^{(l)} \left( \mathbf{h}_v^{(l)}, \; \bigoplus_{u \in \mathcal{N}(v)} \psi^{(l)} \left( \mathbf{h}_v^{(l)}, \mathbf{h}_u^{(l)}, \mathbf{e}_{uv} \right) \right),
\end{equation}

where $\mathcal{N}(v)$ denotes the neighborhood of node $v$, $\psi^{(l)}(\cdot)$ is a message function, $\bigoplus$ is a permutation-invariant aggregation operator (e.g., sum, mean, or max), and $\phi^{(l)}(\cdot)$ is an update function.

A widely used instantiation is the Graph Convolutional Network (GCN) \cite{kipf2016semi}, which performs layer-wise propagation as:

\begin{equation}
\mathbf{H}^{(l+1)} = \sigma \left( \tilde{D}^{-1/2} \tilde{A} \tilde{D}^{-1/2} \mathbf{H}^{(l)} \mathbf{W}^{(l)} \right),
\end{equation}

where $\tilde{A} = A + I$ is the adjacency matrix with self-loops, $\tilde{D}$ is the corresponding degree matrix, $\mathbf{W}^{(l)}$ is a learnable weight matrix, and $\sigma(\cdot)$ is a nonlinear activation function.

Several variants of GNNs have been proposed to address different challenges. For example, GraphSAGE \cite{hamilton2017inductive} introduces neighborhood sampling and aggregation strategies to improve scalability. Simplified Graph Convolution (SGC) \cite{wu2019simplifying} reduces computational complexity by removing nonlinearities and collapsing weight matrices. Early formulations of GNNs were introduced in \cite{scarselli2008graph}, which established the foundation for learning over graph-structured data.

More recent approaches explore Transformer-based architectures for graph learning. Models such as Graphormer \cite{ying2021transformers} leverage attention mechanisms to capture long-range dependencies and global structural information by treating nodes and edges as tokens.

Despite their effectiveness in modeling structured data, GNNs have notable limitations. They often suffer from issues such as over-smoothing, where node representations become indistinguishable across layers, and limited generalization across different graph structures. Furthermore, GNNs typically lack the rich semantic understanding and language reasoning capabilities of large language models when dealing with unstructured textual data.

Overall, GNNs provide a powerful framework for encoding explicit relational structure, making them highly complementary to large language models. While GNNs excel at capturing topology and structured dependencies, LLMs contribute semantic understanding and flexible reasoning over unstructured inputs. This complementarity forms the foundation for graph-LLM integration explored in the subsequent sections.

\section{LLM-Assisted Graph Construction}
This section surveys recent approaches that harness large language models (LLMs) to construct graph-structured representations from unstructured or semi-structured data. As graph-based methods rely on explicit entities and relationships, constructing high-quality graphs is a critical prerequisite for effective graph-LLM integration. Traditional pipelines for graph construction typically involve multiple stages, such as entity recognition, relation extraction, and schema alignment, which often require task-specific models and extensive manual effort.

Recent advances in LLMs have significantly simplified this process by enabling end-to-end or semi-automated graph construction through natural language prompting and structured generation. These approaches allow LLMs to jointly extract entities, infer relationships, and align outputs with predefined or dynamically induced schemas, thereby improving scalability and adaptability across domains.

\begin{figure*}
    \centering
    \includegraphics[width=0.88\linewidth]{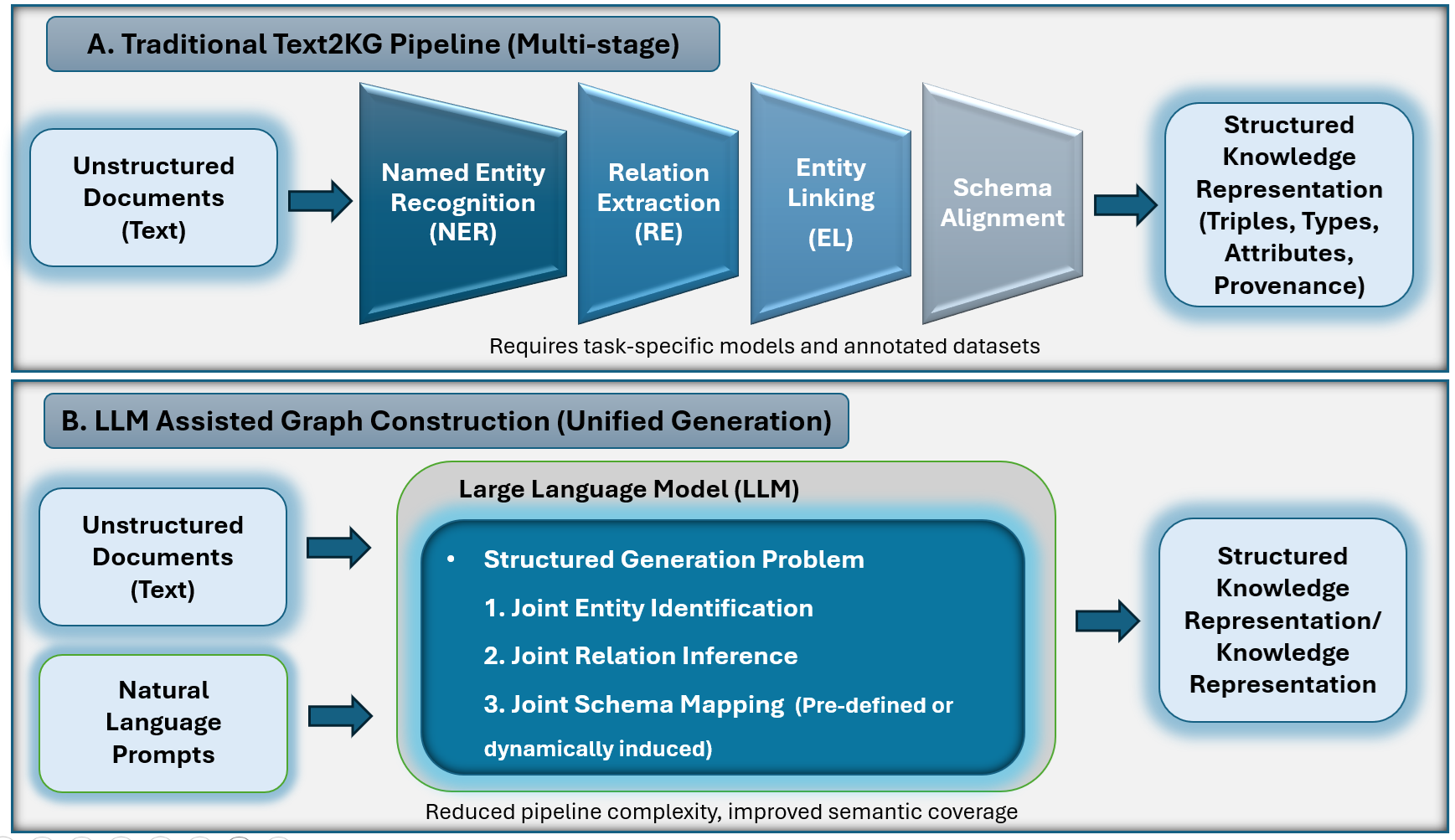}
    \caption{Comparison between traditional multi-stage Text2KG pipelines and LLM-assisted unified graph construction}
    \label{fig:llm_graph_construction}
\end{figure*}

In this section, we review key paradigms for LLM-assisted graph construction, including knowledge graph extraction from text, prompt-based and hybrid extraction pipelines, ontology and schema engineering, and iterative or zero-shot graph construction methods. We also discuss the strengths, limitations, and practical considerations associated with these approaches.
Fig.~\ref{fig:llm_graph_construction} illustrates the difference between traditional multi-stage pipelines and LLM-assisted unified graph construction.

\subsection{Knowledge Graph Extraction from Text}
\label{subsec:text2kg_llm}

Knowledge graph (KG) extraction from text, often referred to as Text2KG, aims to transform unstructured or semi-structured documents into structured knowledge representations, typically in the form of entity-relation-entity triples enriched with types, attributes, and provenance \cite{mihindukulasooriya2023text2kgbench}. Traditional Text2KG pipelines decompose this task into multiple components, including named entity recognition, relation extraction, entity linking, and schema alignment. While effective, such pipelines require task-specific models, annotated datasets, and substantial engineering effort, limiting their scalability and adaptability to new domains.

Recent advances in large language models (LLMs) have significantly reshaped the Text2KG landscape by enabling \emph{LLM-assisted graph construction}, where extraction is formulated as a structured generation problem \cite{sun2025docs2kg}. Instead of relying on separate extraction modules, LLMs can jointly identify entities, infer relations, and map outputs to a predefined or dynamically induced schema using natural language prompts. This paradigm substantially reduces pipeline complexity while improving semantic coverage, particularly for implicit relations and higher-level abstractions expressed in text \cite{gillani2024knowledge, trajanoska2023enhancing, huang2024ctikg}.

\subsection{LLM-assisted extraction paradigms.}
Current research converges on several dominant paradigms for LLM-based KG construction. Prompt-based extraction instructs the LLM to output entities and relations in a constrained format such as JSON or RDF-like triples, often conditioned on a fixed ontology or relation inventory to control relation drift. Hybrid pipelines combine LLM-generated candidates with deterministic validation steps, including schema constraints, duplicate removal, and entity canonicalization, to improve precision and consistency. Retrieval-augmented approaches further ground extraction by incorporating relevant schema definitions, exemplar triples, or external knowledge retrieved at inference time, thereby mitigating hallucinations. More advanced systems adopt agentic or multi-step workflows, decomposing KG construction into extraction, verification, normalization, and consolidation stages with iterative refinement.

\subsection{Ontology Engineering and Schema Construction}
Ontology engineering is another method for connecting LLMs and graphs. For example, in \cite{kommineni2024human}, the authors focused on automating ontology and knowledge graph construction using large language models (LLMs) to reduce reliance on human experts. Their goal was to design a semi-automated pipeline that generates competency questions, builds an ontology schema, extracts entities and relationships from scientific texts, and constructs a knowledge graph with minimal manual intervention. They applied retrieval-augmented generation (RAG) and tested multiple open-source LLMs on a dataset of scientific publications related to deep learning. Their results showed that LLMs can effectively support ontology and knowledge graph creation with reasonable accuracy.

In \cite{meyer2023llm}, the authors of the paper focused on exploring how large language models like ChatGPT can assist with knowledge graph engineering (KGE), especially to support tasks that normally require deep expertise in graph structures, vocabularies, logic, and web technologies. They conducted comprehensive experiments to test whether ChatGPT (including GPT-3.5 and GPT-4) can help with the development and management of knowledge graphs by performing specific engineering tasks and evaluated the results of these experiments to demonstrate the model’s potential and limitations in supporting KGE processes.

In \cite{barua2022complex}, the authors focused on applying large language models (LLMs) to the challenging task of complex ontology alignment, which means finding semantic correspondences between detailed, multi-entity structures in different ontologies rather than just simple one-to-one matches. Their goal was to replicate and evaluate a previous approach that uses modular information about ontologies to guide LLMs in generating alignment rules, testing whether this method remains effective on a new dataset with more varied, complex alignments. They used a dataset based on the Enslaved ontology to propose alignment rules.

\subsection{Iterative and Zero-Shot Graph Construction}
Also, other researchers focused on zero-shot graph construction. In \cite{carta2023iterative}, the authors focused on automatically constructing knowledge graphs using large language models (LLMs) in a zero-shot setting, without relying on annotated data or predefined ontologies. Their goal was to design a scalable and general pipeline that extracts entities, relations, and triplets directly from unstructured text through iterative prompting. Using models like GPT-3.5, they demonstrated that LLMs can generate structured knowledge graphs effectively, showing the potential of prompt-based approaches for automated knowledge graph construction.

In \cite{sansford2024grapheval}, the authors focused on creating a new method to evaluate and detect hallucinations (inaccurate or inconsistent outputs) from large language models by converting their responses into knowledge graph (KG) structures and checking each part against known context. Their goal was to provide a more explainable and systematic way to pinpoint exactly where a model’s answer diverges from factual grounding than previous metrics, while also improving detection accuracy by combining their KG representation with natural language inference models. They also introduced a follow-up method called GraphCorrect that leverages the same KG structure to attempt to correct hallucinated information.

In \cite{meyer2023developing}, the authors focused on creating a scalable benchmark called LLM-KG-Bench to assess how well large language models perform on tasks related to knowledge graph engineering (KGE). Their goal was to design a framework with multiple challenge tasks, including syntax/error correction, fact extraction, and dataset generation, that can automatically evaluate LLM responses, track prompt engineering, and visualize performance, because existing benchmarks didn’t adequately measure LLMs’ abilities in KGE. They found that while LLMs can be useful tools, they are not yet effective enough for knowledge graph generation using zero-shot prompting, and their benchmark helps quantify and compare model strengths and weaknesses in this area.

\section{Graph-Enhanced LLM Reasoning}
\begin{figure*}
    \centering
    \includegraphics[width=0.99\linewidth]{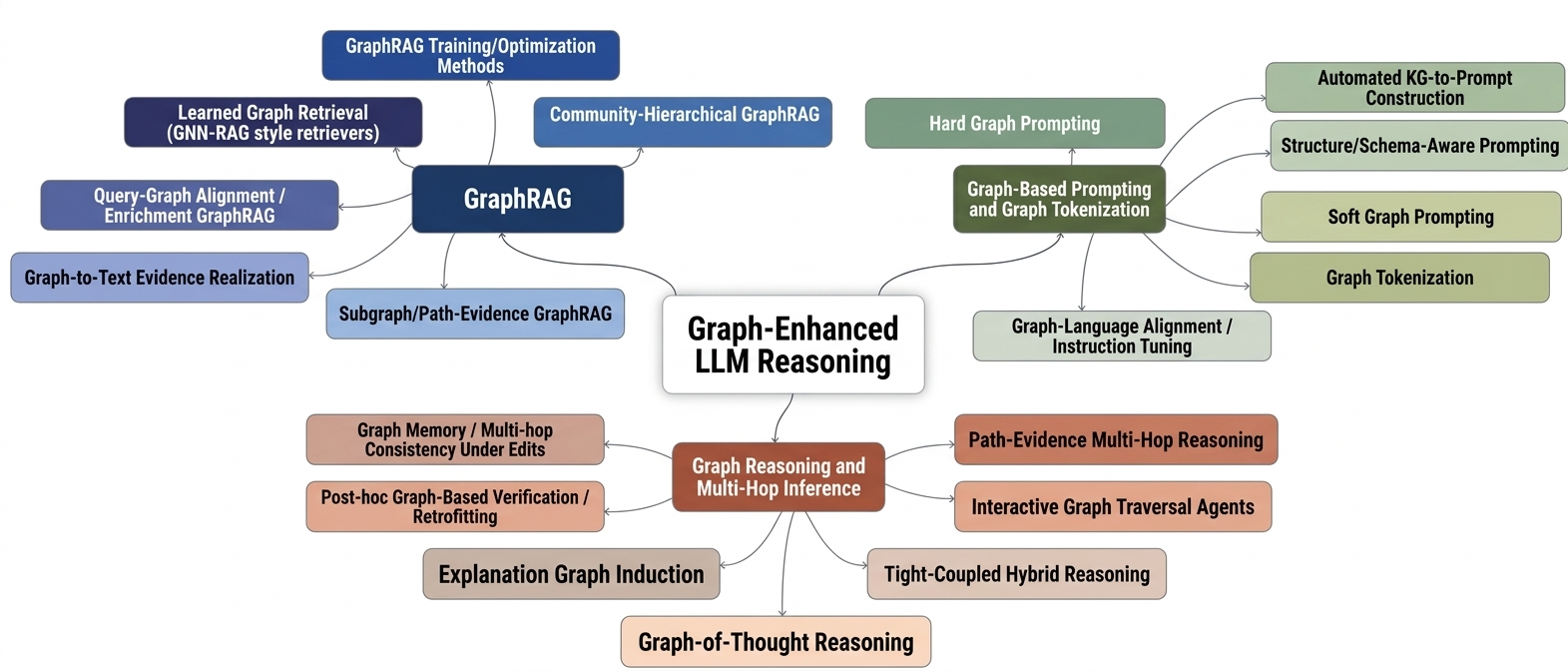}
    \caption{Overview of graph-enhanced LLM reasoning paradigms, including GraphRAG, graph prompting, and graph-guided inference methods}
    \label{fig:graph_reasoning}
\end{figure*}

This section surveys graph-enhanced LLM reasoning paradigms, where graph structures are incorporated directly into the inference pipeline to impose relational constraints, enable multi-hop reasoning, and improve factual consistency. We review Graph Retrieval-Augmented Generation (Graph-RAG) methods that perform structure-aware retrieval over graphs (Section 4.1), graph-based prompting and tokenization techniques that encode topology and relational semantics into LLM inputs or representations (Section 4.2), and graph-guided reasoning frameworks in which explicit graph traversal or path selection steers LLM inference (Section 4.3). Fig.~\ref{fig:graph_reasoning} shows a general view of LLM-graph reasoning methods.

\subsection{Graph Retrieval-Augmented Generation (RAG)}
In a Direct LLM workflow, the model receives a user query and produces an answer solely based on its internal parametric knowledge. All factual recall, reasoning, and multi-step inference are implicitly encoded in the model’s learned weights. While this approach offers a simple and efficient inference pipeline, it exhibits fundamental limitations. The model’s knowledge is static and frozen at training time, performance degrades on domain-specific or long-tail queries, and multi-hop reasoning remains unreliable because intermediate relational constraints are not explicitly represented or enforced \cite{wu2025medical}. Consequently, Direct LLMs are prone to hallucination, brittle reasoning chains, and limited transparency, particularly when answering questions that require integrating multiple interdependent facts.
Retrieval-Augmented Generation (RAG) addresses the knowledge limitation of Direct LLMs by introducing an external retrieval step. Given a query, relevant documents or text chunks are retrieved from a corpus—typically using vector similarity—and supplied to the LLM as contextual grounding before generation. This design reduces hallucination by anchoring responses in external evidence and enables access to up-to-date or domain-specific information \cite{hu2025self}. However, classical RAG treats retrieved content as a flat collection of independent passages, largely ignoring structural relationships among entities, facts, or documents. As a result, while RAG improves factual recall, it continues to struggle with multi-hop reasoning, compositional queries, and global logical consistency. In complex reasoning scenarios, relevant pieces of information may be retrieved, yet the model lacks an explicit mechanism to coherently chain them.
\begin{table*}[t]
\centering
\caption{Comparison of Direct LLM, RAG, and GraphRAG paradigms in terms of workflow, strengths, and limitations}
\label{tab:llm_rag_graphrag_comparison}
\renewcommand{\arraystretch}{1.15}
\begin{tabularx}{\textwidth}{l X X X}
\toprule
\textbf{Paradigm} & \textbf{Core Workflow} & \textbf{Strengths} & \textbf{Weaknesses} \\
\midrule
\textbf{Direct LLM} &
Query processed directly by LLM for answer generation &
Static inference, low latency, strong language fluency &
Static knowledge, hallucinations, weak multi-hop reasoning \\
\midrule
\textbf{RAG} &
Retrieve relevant documents and inject contextual evidence before LLM generation &
External grounding, reduced hallucination, updatable corpus &
Flat retrieval, no relational structure, weak compositional reasoning \\
\midrule
\textbf{GraphRAG} &
Graph-aware retrieval followed by subgraph reasoning with LLM &
Structure-aware retrieval, explicit multi-hop reasoning, interpretability &
Graph construction cost, higher system complexity, scalability challenges \\
\bottomrule
\end{tabularx}
\end{table*}

Graph Retrieval-Augmented Generation (GraphRAG) extends the RAG paradigm by explicitly modeling retrieved knowledge as a graph, where nodes represent entities, documents, or concepts, and edges encode semantic, temporal, causal, or relational dependencies. Rather than retrieving isolated text chunks, GraphRAG retrieves subgraphs or reasoning paths, preserving the relational structure required for multi-step inference \cite{peng2025graph, mavromatis2025gnn, li2025enrich}. This enables controlled traversal, neighborhood expansion, and constraint-aware reasoning, allowing the LLM to follow explicit relational pathways instead of implicitly inferring them from unstructured text. 
\\
In summary, GraphRAG can be viewed as a natural evolution of retrieval-augmented generation, where grounding alone is insufficient and explicit relational structure is required to support reliable, interpretable, and generalizable reasoning by large language models. The key differences in workflow design, strengths, and limitations among Direct LLMs, RAG, and GraphRAG are summarized in Table~\ref{tab:llm_rag_graphrag_comparison}.


\section{Hybrid GNN-LLM Models}

Recent advances in integrating Large Language Models (LLMs) with Graph Neural Networks (GNNs) have led to a rapidly evolving research landscape. Existing works can be broadly categorized into four paradigms based on their interaction mechanisms and architectural designs: (1) collaboration frameworks, (2) directional integrations, (3) explainability-enhanced hybrids, and (4) pre-trained hybrid architectures. This taxonomy is consistent with recent surveys that categorize LLM-graph integration based on coupling strategies between language and graph models~\cite{ren2024survey}.

\begin{figure*}[t]
    \centering
    \includegraphics[width=0.99\linewidth]{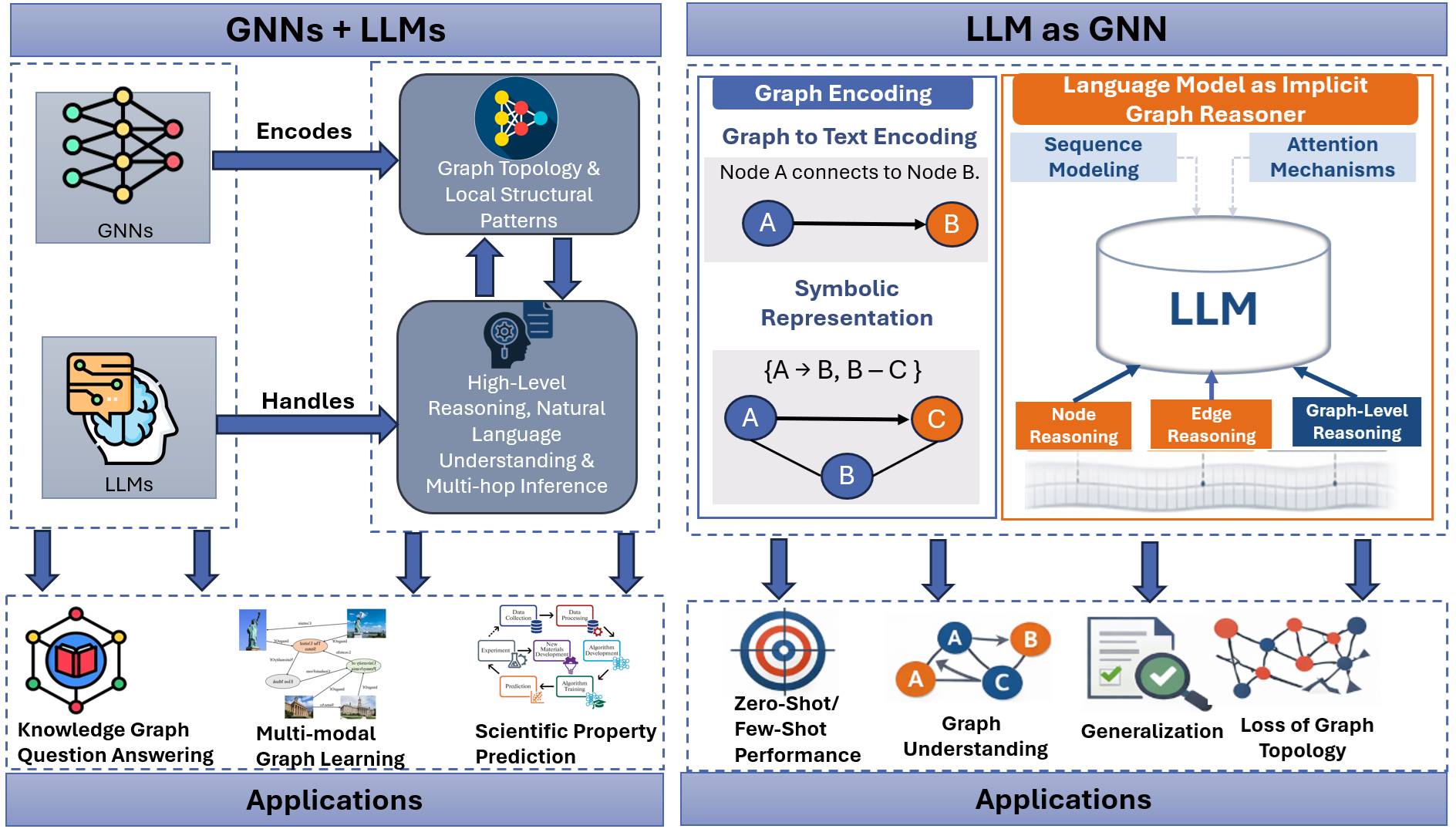}
    \caption{Conceptual comparison of hybrid GNN–LLM frameworks and the LLM-as-GNN paradigm, highlighting differences in structural and semantic integration}
    \label{fig:hybrid_gnn_llm}
\end{figure*}

Fig.~\ref{fig:hybrid_gnn_llm} provides a conceptual overview of these paradigms by contrasting hybrid GNN-LLM collaboration frameworks with the emerging LLM-as-GNN paradigm. Hybrid approaches explicitly combine structural inductive bias with semantic reasoning, whereas LLM-as-GNN methods implicitly encode graph structure into textual sequences and rely on attention mechanisms for reasoning.

A unified comparison of representative methods across all paradigms is provided in Table~\ref{tab:global_comparison}.

\subsection{Collaboration Frameworks}

Collaboration frameworks enable bidirectional interaction between LLMs and GNNs through iterative or cooperative mechanisms. These approaches can be categorized into:
(1) iterative co-training, 
(2) LLM-guided supervision, and 
(3) graph structure refinement.

\textbf{Iterative Co-training:}Iterative co-training frameworks establish feedback loops between LLMs and GNNs. Representative methods such as GLEM~\cite{zhao2022learning} and LOGIN~\cite{qiao2025login} alternately refine graph representations and predictions using both semantic and structural signals.

\textbf{LLM-Guided Supervision} : In this setting, LLMs provide pseudo-labels, rationales, or soft supervision signals. Distillation-based approaches~\cite{pan2024distilling} enable knowledge transfer from LLMs to GNNs, improving performance in low-resource settings.

\textbf{Graph Structure Refinement} : These approaches leverage LLMs to modify graph topology. GraphEdit~\cite{guo2024graphedit} is a representative example that refines graph connectivity to improve downstream learning.

\subsection{Directional Integrations: LLM4GNN and GNN4LLM}

Directional integration represents unidirectional interaction, where either LLMs enhance graph learning (LLM4GNN) or graph structures enhance LLM reasoning (GNN4LLM).

\subsubsection{LLM4GNN}

LLM4GNN approaches can be divided into:
(1) feature augmentation, 
(2) pseudo-labeling and supervision, and 
(3) graph structure enhancement.

Feature augmentation methods (e.g., GLEM~\cite{zhao2022learning}, LLaGA~\cite{chen2024llaga}) improve node representations using LLM embeddings. Pseudo-labeling approaches (e.g., LOGIN~\cite{qiao2025login}, distillation~\cite{pan2024distilling}) provide supervision signals. Structure enhancement methods (e.g.,~\cite{zhang2025can}, SaVe-TAG~\cite{wang2024save}) refine graph topology or generate data.

\subsubsection{GNN4LLM}

GNN4LLM approaches can be categorized into:
(1) graph retrieval, 
(2) graph-guided prompting, and 
(3) instruction-tuned reasoning. GNN-RAG~\cite{mavromatis2024gnn} enables graph-based retrieval for reasoning, while LLaGA~\cite{chen2024llaga} supports graph-to-text encoding. InstructGraph~\cite{wang2024instructgraph} enables graph-aware instruction tuning for zero-shot reasoning.

\subsection{Explainability-Enhanced Hybrid Models}

These models aim to improve interpretability by combining graph reasoning with natural language explanations.

They can be categorized into:
(1) natural language explanation generation,
(2) rationale-guided learning, and
(3) structure-aware explanation modeling.

Gspell~\cite{baghershahi2025gspell} generates both textual explanations and supporting subgraphs. Distillation approaches~\cite{pan2024distilling} use rationales as supervision signals. Structure-aware models such as LLaGA~\cite{chen2024llaga} ensure explanations are grounded in graph topology.

\subsection{Pre-trained Hybrid Architectures}

Pre-trained hybrid architectures aim to learn general-purpose representations across graph and text modalities.

These approaches can be categorized into:
(1) instruction-tuned graph LLMs,
(2) graph-language co-pretraining, and
(3) graph-native foundation models.

Instruction-tuned models (GraphGPT~\cite{tang2024graphgpt}, InstructGraph~\cite{wang2024instructgraph}, HiGPT~\cite{tang2024higpt}) adapt LLMs for graph reasoning tasks. Co-pretraining approaches (LLaGA~\cite{chen2024llaga}) align graph and text representations. Graph-native models (GOFA~\cite{kong2024gofa}, GDL4LLM~\cite{zhou2025each}) integrate graph computation directly into LLM architectures.

\begin{table*}[t]
\centering
\caption{Global comparison of representative GNN-LLM hybrid models across paradigms.}
\label{tab:global_comparison}
\resizebox{\textwidth}{!}{%
\begin{tabular}{l l l l l l l}
\toprule
\textbf{Method} & \textbf{Year} & \textbf{Paradigm} & \textbf{Subtype} & \textbf{Interaction} & \textbf{Key Mechanism} & \textbf{Strength} \\
\midrule
GLEM~\cite{zhao2022learning} & 2022 & Collaboration / LLM4GNN & Co-training & Bidirectional & EM-style optimization & Strong graph-text alignment \\
LOGIN~\cite{qiao2025login} & 2024 & Collaboration / LLM4GNN & Co-training & Iterative & LLM as consultant & Selective knowledge injection \\
Distillation~\cite{pan2024distilling} & 2024 & Collaboration / Explainability & Supervision & One-way & Knowledge transfer & Efficient learning \\
GraphEdit~\cite{guo2024graphedit} & 2024 & Collaboration & Structure refinement & Indirect & Graph editing via LLM & Improved graph quality \\
\midrule
Robustness~\cite{zhang2025can} & 2025 & LLM4GNN & Structure & One-way & Edge refinement & Robust learning \\
SaVe-TAG~\cite{wang2024save} & 2024 & LLM4GNN & Augmentation & One-way & Data generation & Handles long-tail data \\
GNN-RAG~\cite{mavromatis2024gnn} & 2024 & GNN4LLM & Retrieval & One-way & Graph retrieval & Multi-hop reasoning \\
LLaGA~\cite{chen2024llaga} & 2024 & GNN4LLM / Explainability & Prompting & One-way & Graph-to-text encoding & Structured reasoning \\
InstructGraph~\cite{wang2024instructgraph} & 2024 & GNN4LLM / Pre-trained & Instruction & One-way & Instruction tuning & Zero-shot ability \\
\midrule
Gspell~\cite{baghershahi2025gspell} & 2025 & Explainability & NL + Structure & Hybrid & Text + subgraph explanation & Interpretability \\
\midrule
GraphGPT~\cite{tang2024graphgpt} & 2023 & Pre-trained & Instruction & Unified & Graph instruction tuning & Generalization \\
HiGPT~\cite{tang2024higpt} & 2024 & Pre-trained & Instruction & Unified & Graph tokenization & Heterogeneous graphs \\
GOFA~\cite{kong2024gofa} & 2024 & Pre-trained & Graph-native & Unified & GNN inside LLM & Deep integration \\
GDL4LLM~\cite{zhou2025each} & 2025 & Pre-trained & Graph-native & Unified & Graph-as-language & Efficient encoding \\
\bottomrule
\end{tabular}%
}
\end{table*}

\section{Knowledge Graph Question Answering (KGQA)}
This section surveys recent advances in knowledge graph question answering (KGQA), a key application domain of graph-LLM integration. KGQA aims to answer natural language queries by reasoning over structured knowledge graphs, where entities and relationships are explicitly represented. By grounding reasoning in graph structures, KGQA systems enable more interpretable and multi-hop inference compared to purely text-based approaches.

Traditional KGQA methods rely on semantic parsing, symbolic reasoning, or supervised learning over structured queries, often requiring large annotated datasets and carefully designed pipelines. However, these approaches can struggle with scalability, domain adaptation, and handling complex or ambiguous queries.

Recent developments in large language models (LLMs) have significantly reshaped the KGQA landscape. LLMs can act as reasoning engines, semantic parsers, or agents that guide graph traversal and inference, enabling more flexible and generalizable question answering. At the same time, integrating LLMs with knowledge graphs helps mitigate hallucinations and improves factual grounding by constraining generation with structured evidence.

In this section, we review both training-free and LLM-augmented KGQA frameworks, highlighting their design principles, strengths, and limitations. We also discuss how different integration strategies balance symbolic reasoning with generative capabilities to support accurate and scalable question answering over knowledge graphs.
\subsection{Training-Free and Low-Resource KGQA}
Training-free and low-resource KGQA methods aim to reduce reliance on large annotated datasets and costly fine-tuning procedures. These approaches leverage prompt-based reasoning, symbolic graph traversal, or lightweight adaptation techniques to enable question answering over knowledge graphs with minimal supervision, making them suitable for domain-specific or rapidly evolving knowledge graphs.

Representative systems demonstrate that LLMs can perform dynamic reasoning, link prediction, and multi-hop inference without full retraining. Although their performance may not always match fully supervised models, these methods significantly lower deployment barriers and highlight the practicality of LLM-based KGQA in data-scarce settings.

In \cite{tao2024finqa}, the authors focus on building a training-free question answering system over a dynamic knowledge graph tailored to the financial domain. They propose FinQA, which constructs a dynamic finance knowledge graph partitioned by update frequency, designs a training-free pipeline that parses natural language questions into graph query language (NL2GQL), and integrates an open-source large language model for revision to improve parsing accuracy. The method targets the challenge of frequent data updates and scarce annotated training data in financial KGQA and evaluated on real-world finance KGQA tasks, demonstrating strong performance in dynamic finance question answering.

In \cite{zhang2024gail}, the authors focus on improving knowledge graph question answering (KGQA) in low-resource settings by enhancing large language models (LLMs) through fine-tuning with generative adversarial imitation learning (GAIL). They propose a framework that uses GAIL to guide the LLM’s reasoning behavior, aiming to generate more accurate answers over a knowledge graph when annotated training data is scarce. The method is evaluated on benchmark KGQA datasets to demonstrate its effectiveness in low-resource QA scenarios.

In \cite{shu2024knowledge}, the authors focus on improving multi-hop link prediction in knowledge graphs by introducing a framework called KG-LLM, which leverages large language models (LLMs) to convert structured graph data into natural language prompts and fine-tune the LLMs to reason over graph connections. They transform graph paths into chain-of-thought-style natural language prompts and use instruction fine-tuning (and optionally in-context learning) to train models such as Flan-T5, LLaMA2, and Gemma to predict whether distant entities are connected. The method is evaluated on standard multi-hop link prediction benchmark datasets such as WN18RR and NELL-995, demonstrating improved generalization and prediction accuracy in unfamiliar scenarios

\subsection{LLM-Augmented KGQA Frameworks}
LLM-augmented KGQA frameworks integrate large language models into the reasoning pipeline to enhance semantic parsing, candidate answer generation, and logical inference. In these systems, LLMs often act as reasoning agents that guide graph traversal, generate executable queries, or compensate for incompleteness and noise in knowledge graphs.

Experimental results indicate that LLM augmentation improves robustness and reasoning accuracy, particularly for complex, multi-hop, and domain-specific queries. Nevertheless, challenges remain in mitigating hallucination, ensuring faithfulness to graph evidence, and balancing generative flexibility with symbolic correctness, motivating ongoing research on tighter integration and verification mechanisms.
In this  \cite{an2023knowledge}, the authors focus on designing a knowledge-graph question answering system tailored to materials science. They introduce KGQA4MAT, a framework that leverages knowledge graphs constructed for material entities and properties, coupled with pre-trained large language models to interpret and answer natural language questions related to materials data. The method combines structured graph retrieval with semantic language understanding to support complex queries over materials knowledge graphs, and the approach is evaluated on domain-specific KGQA tasks involving materials databases and related benchmarks.
 In this \cite{xu2024generate}, the authors focus on addressing question answering over incomplete knowledge graphs, where the knowledge graph does not contain all the triples needed to answer a question. They propose a training-free method called Generate-on-Graph (GoG) that treats a large language model (LLM) both as an agent that searches the graph and as a knowledge source that generates new factual triples to augment the graph. GoG uses a Thinking-Searching-Generating framework, the model iteratively explores the graph, identifies missing information, and synthesizes additional facts to enable correct answers. They evaluate their approach on two IKGQA (incomplete KG Question Answering) datasets constructed from existing KGQA benchmarks and show that GoG outperforms previous methods in this setting

In this \cite{xu2024llm}, the authors focus on improving large language model (LLM) performance on knowledge graph question answering (KGQA) by reducing hallucinated or ungrounded reasoning results produced by generative LLMs. They propose a framework called READS that reformulates KGQA into a series of discriminative subtasks (including subgraph search, subgraph pruning, and answer inference) and designs a corresponding discriminative inference strategy to better guide reasoning over the knowledge graph. The approach is evaluated on widely used benchmark KGQA datasets, such as WebQSP and Complex Web Questions (CWQ), and achieves state-of-the-art performance compared with strong baselines


\section{Scene Graphs and Large Language Models}
\begin{figure*}[h]
    \centering
    \includegraphics[width=\linewidth]{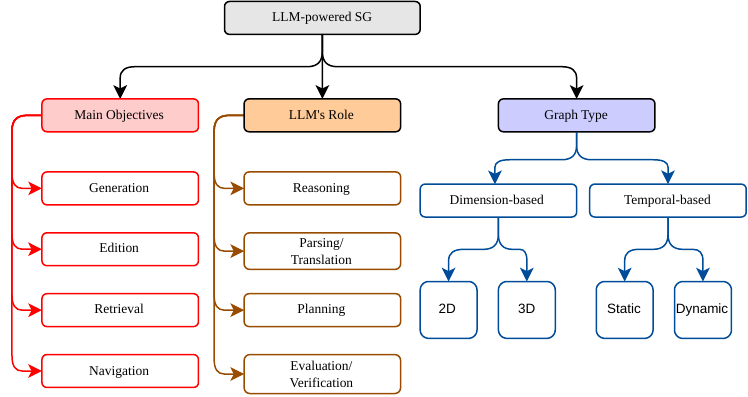}
    \caption{Taxonomy of LLM-enhanced scene graph frameworks based on objectives, LLM roles, and graph types.}
    \label{fig:llm-enhanced-sg-taxonomy}
\end{figure*}

Scene graphs (SGs) represent objects, relationships, and attributes existing in a scene \cite{li2024scene,chang2021comprehensive,chang2021scene}. LLMs can be used as advanced tools to facilitate scene-graph-centric tasks such as generation, editing, and navigation by playing different roles across the pipeline. In the following sections, we explain the different dimensions of the integration of LLM and SGs as shown in Fig.~\ref{fig:llm-enhanced-sg-taxonomy}.

\subsection{Graph Type}
The graph type is SGs can be viewed from spatial dimensionality and temporal setting. Based on spatial dimensionality, the graph could either be 2D or 3D. A 2D scene graph models relations in the image plane, while a 3D scene graph additionally encodes metric depth and object geometry, enabling physically grounded spatial reasoning \cite{li2024scene,chang2021comprehensive,zemskova20253dgraphllm}. Temporal setting describes whether or not the graph evolves over time. So, based on the temporal setting, the graph could either be static or dynamic. A static graph is constructed from a single observation and represents one fixed scene or time instant, although it may be internally refined during optimization. In contrast, a dynamic graph evolves as new or sequential observations update the scene representation \cite{li2024scene,chang2021comprehensive,dai2024optimal,yin2024sg}.
\subsection{LLM's Role}
LLMs play important roles in SG tasks. These roles include reasoning, parsing/translation, planning, and evaluation/verification.

In reasoning LLMs are used to infer semantics, relationships, or latent attributes, which can't be directly observed \cite{zhang2026scenellm,lv2024sgformer}. In parsing/translation, LLMs are used to directly convert unstructured inputs (text or instructions) into structured presentations such as scene graphs, triplets, logic, or commands \cite{kim2024llm4sgg,huang2024toward,dai2024optimal}. In planning, LLMs are used to make decisions, organize actions, select goals, or guide search \cite{dai2024optimal,colombani2024time,yin2024sg}. In evaluation, LLMs are used to assess correctness, consistency, or alignment between representations \cite{chen2024makes,yang2025llm,zhao2023less,wang2025sakr}. 
\subsection{Main Objectives}
The main objectives of the integration of LLMs and SGs are divided into generation, editing, retrieval, and navigation.
\subsubsection{Generation}
In generation, the primary objective is to create a new structured representation or scene, typically a scene graph or a scene itself, from raw or unstructured inputs.

In \cite{zhang2026scenellm}, the authors propose SceneLLM for dynamic scene graph generation (SGG). The framework consists of three phases: video-to-language (v2l) mapping, reasoning with an LLM, and SG prediction. First, V2L maps the video frames to an implicit linguistic signal (a sequence of scene tokens) by discretizing object features and embedding spatial-temporal information. Then, in the second phase, they embed the linguistic tokens into a prompt and feed it to a LoRA fine-tuned LLM to generate reasoned hidden representations. Finally, in SG prediction, a transformer-based SGG predictor generates the final SG based on the hidden states from the previous step. The results demonstrate the state-of-the-art performance of SceneLLM in terms of \texttt{recall@K} for the Action Genome (AG) dataset across various tasks.

In \cite{hu2024scenecraft}, the authors propose SceneCraft for static-3D SGG by transforming a text query into a 3D scene. This pipeline consists of four phases: asset retrieval and scene decomposition, SG construction, scene layout optimization, and library learning. In the first phase, an LLM makes a list of assets (3D objects) and their descriptions from a text query. Then for each asset the top-10 3D models are retrieved from a large depository and the one with the highest text-to-image score is kept as final. Then, SceneCraft breaks crowded scenes into smaller sub-scenes to solve the problem more efficiently. In the second phase, a layout matrix is constructed for every asset to put each of them in the correct location and orientation. This done by constructing a relational scene graph with help of LLM-Planner. In the third phase, SceneCraft uses a set of scoring functions to iteratively optimize the scene layout using a Multimodal LLM's feedback to obtain the final optimized scene.
\subsubsection{Editing}
In addition, the goal of LLM is to modify an existing scene or scene graph to obtain a new configuration.

EditRoom \cite{zheng2024editroom} is a static-3D \underline{s}cene \underline{g}raph \underline{e}dition (SGE) framework, which is capable of six types of edits: rotate, translate, scale, replace, add, and remove. EditRoom receives a natural language text and the original scene as input to make the final scene; this includes two main modules: command parameterizer and scene editor. In the command parameterizer, the natural language command is converted into a set of edit types. After that, EditRoom uses graph diffusion models to find a target scene with the highest probability.

\underline{S}cene \underline{g}raph \underline{e}dit (SGEdit) uses parsing and planning capabilities of LLMs to perform static-2D SGE tasks. SGEdit consists of two main stages: scene parsing and image editing. Given an input image, SGEdit generates a scene graph, masks, and descriptions using an LLM-driven scene parser. After that, a diffusion model is fine-tuned to learn each object's identity. In the next stage, the LLM acts as a planner to decide the changes and how to apply them, and a diffusion model performs editing tasks.
\subsubsection{Retrieval}
In retrieval, scene graphs are used as structured indices to search, match, rank, or query information, rather than to generate or modify content.

In \cite{jeong2025llm}, the authors propose \underline{V}isual \underline{T}riplet-based \underline{G}raph \underline{T}ransformation (VTGT) 
for static-2D object retrieval from input images. The pipeline consists of two phases: graph transformation and graph representation learning. In phase one, VTGT transforms a scene graph into a triplet graph, where nodes include subject-relation-object triplets and edges represent similarity between the nodes; this is achieved by introducing local and global features for each triplet. Local features are used to keep the ordered sequence, while global features are attained by help of an LLM. The encodings for these two feature types are then merged to be used in the next phase. In the second phase, a \underline{g}raph \underline{a}ttention \underline{n}etwork is used to update the representation of each triplet based on its top-k similar triplets. Finally, a single semantic embedding for the input image is generated, which is used for downstream retrieval tasks.

3DGraphLLM \cite{zemskova20253dgraphllm} is a static-3D method that takes point clouds of scene objects and builds a new scene representation to support object retrieval from natural language queries. It assigns each object both 2D features (appearance cues such as texture and color) and 3D features (shape and geometry). Then, semantic relations between object pairs are extracted using a vision-language spatial attention transformer and encoded as latent relation vectors. Finally, the objects and their relations are projected into the LLM token space and fed to an LLM (via object identifier tokens and a compact k-nearest-neighbor subgraph representation) to retrieve the target object from the scene.
\subsubsection{Navigation}
In navigation, scene graphs are used to guide an embodied agent’s movement or task execution within an environment.

In \cite{colombani2024time}, the authors propose \underline{T}ime \underline{i}s \underline{i}n \underline{m}y \underline{S}ight (TioMS), which is an LLM-driven robot architecture for dynamic environments using continuously updated scene graphs. The pipeline has two main modules: perception and planning. In the perception module, RGB-D frames are processed by a scene graph generator to detect objects and relationships, then object positions are estimated in 3D using robot/camera pose, and a particle filter refines tracking and localization across frames before updating the semantic map. In the planning module, the LLM uses the updated semantic map to translate natural-language commands into robot skills and replans when execution fails.

In \cite{yin2024sg}, the authors propose \underline{s}cene \underline{g}raph \underline{nav}igation (SG-Nav) for zero-shot object-goal navigation by using an online 3D scene graph as the main representation for LLM-based planning. This pipeline consists of three main steps: online scene graph construction, LLM-guided exploration, and re-perception. In the first step, the agent incrementally builds a hierarchical 3D scene graph from RGB-D observations and prunes unreliable edges using geometric constraints and VLM-based checks. In the second step, the LLM scores graph regions to select promising exploration frontiers for searching the target object. In the third step, when the target object is detected, SG-Nav re-observes it from multiple views to confirm the detection and reduce false positives before finalizing the navigation result.


A detailed comparison of state-of-the-art LLM-powered scene graph frameworks is presented in Table~\ref{tab:sg-comprehensive-table}.

\begin{table*}[t]
\centering
\caption{Comprehensive comparison of state-of-the-art LLM-powered scene graph frameworks across roles and graph types (R1: reasoning, R2: parsing/translation, R3: planning, R4: evaluation/verification).}
\label{tab:sg-comprehensive-table}
\resizebox{\textwidth}{!}{
\begin{tabular}{l c c c c c c c c}
\toprule
\textbf{Research Work} & \textbf{Year} & \textbf{Objective} & \textbf{Input} & \multicolumn{4}{c}{\textbf{LLM Role}} & \textbf{Graph Type} \\
\cmidrule(lr){5-8}
 &  &  &  & \textbf{R1} & \textbf{R2} & \textbf{R3} & \textbf{R4} &  \\
\midrule

SceneLLM \cite{zhang2026scenellm} & 2026 & Generation & Video & \checkmark & \checkmark & \ding{55} & \ding{55} & Dynamic-2D \\
SceneCraft \cite{hu2024scenecraft} & 2024 & Generation & Text & \checkmark & \checkmark & \checkmark & \checkmark & Static-3D \\
LLM4SGG \cite{kim2024llm4sgg} & 2024 & Generation & Image+Text & \checkmark & \checkmark & \ding{55} & \ding{55} & Static-2D \\
SDSGG \cite{chen2024scene} & 2024 & Generation & Image & \checkmark & \ding{55} & \ding{55} & \ding{55} & Static-2D \\
PASGG-LM \cite{platnick2024enabling} & 2024 & Generation & Image & \checkmark & \ding{55} & \ding{55} & \ding{55} & Static-2D \\
Planner3D \cite{wei2025planner3d} & 2025 & Generation & Graph/Text & \checkmark & \ding{55} & \ding{55} & \ding{55} & Static-3D \\
LLaVA-SpaceSGG \cite{xu2025llava} & 2025 & Generation & Image & \checkmark & \checkmark & \ding{55} & \ding{55} & Static-2D \\
GraLa3D \cite{huang2024toward} & 2024 & Generation & Text & \ding{55} & \checkmark & \checkmark & \ding{55} & Static-3D \\
IntegraPSG \cite{zhao2025integrapsg} & 2025 & Generation & Image & \checkmark & \ding{55} & \ding{55} & \ding{55} & Static-2D \\
SGFormer \cite{lv2024sgformer} & 2024 & Generation & Point cloud & \checkmark & \ding{55} & \ding{55} & \ding{55} & Static-3D \\
ELEGANT \cite{zhao2023less} & 2023 & Generation & Image & \checkmark & \checkmark & \ding{55} & \checkmark & Static-2D \\

HRSGL \cite{li2024llm} & 2024 & Editing & 3D scene & \checkmark & \ding{55} & \checkmark & \checkmark & Static-3D \\
SGEdit \cite{zhang2024sgedit} & 2024 & Editing & Image & \checkmark & \checkmark & \checkmark & \ding{55} & Static-2D \\
EditRoom \cite{zheng2024editroom} & 2025 & Editing & 3D Scene+Text & \ding{55} & \checkmark & \checkmark & \ding{55} & Static-3D \\
ScanEdit \cite{el2025scanedit} & 2025 & Editing & 3D Scene+Text & \checkmark & \ding{55} & \checkmark & \ding{55} & Static-3D \\
SAKR-Edit \cite{wang2025sakr} & 2025 & Editing & Image+Text & \checkmark & \ding{55} & \ding{55} & \ding{55} & Static-2D \\

VTGT \cite{jeong2025llm} & 2025 & Retrieval & Image & \checkmark & \ding{55} & \ding{55} & \ding{55} & Static-2D \\
3DGraphLLM \cite{zemskova20253dgraphllm} & 2025 & Retrieval & 3D Scene & \checkmark & \ding{55} & \ding{55} & \ding{55} & Static-3D \\

TioMS \cite{colombani2024time} & 2024 & Navigation & Video & \ding{55} & \checkmark & \checkmark & \ding{55} & Dynamic-3D \\
OSGP \cite{dai2024optimal} & 2024 & Navigation & Graph & \ding{55} & \checkmark & \checkmark & \ding{55} & Static-3D \\
SG-Nav \cite{yin2024sg} & 2024 & Navigation & Video & \checkmark & \ding{55} & \checkmark & \ding{55} & Dynamic-3D \\

\bottomrule
\end{tabular}
}
\end{table*}

\section{Graph-Agent-LLM Integration}
The authors in \cite{ji2025lafa} focused on developing LAFA (LLM-Agentic Federated Analytics), a hierarchical multi-agent framework that enables complex natural-language analytics over decentralized data while preserving privacy. Instead of centralizing raw data, their system uses a structured agent pipeline—comprising a coarse-grained planner, a fine-grained planner, a DAG optimizer, and an answerer agent—to translate user queries into optimized execution plans for federated analytics. The framework leverages Large Language Models (LLMs) to decompose and structure analytical tasks, while federated analytics backends handle secure aggregation and privacy mechanisms such as encryption and differential privacy. For evaluation, the authors used the AdultPii dataset (32,563 records with 18 features) and a benchmark of 20 complex natural-language analytics queries generated using GPT-4o based on realistic privacy scenarios (including Apple privacy reports). The results demonstrate improved semantic parsing accuracy, higher execution success rates than those of baseline prompting strategies, and a significant reduction in redundant federated operations through DAG optimization, thereby improving efficiency while maintaining strict privacy guarantees. \\

The authors in \cite{chen2025x} focused on developing X-GridAgent. This LLM-powered agentic AI system automates complex power grid analysis using natural-language queries while integrating domain-specific tools and structured databases for rigorous engineering computations. Their model uses a three-layer hierarchical architecture (planning, coordination, and action layers) in which the planning layer interprets user intent and generates structured workflows, the coordination layer manages task execution and memory, and the action layer interfaces with professional tools (e.g., Pandapower modules) to perform analyses like power flow, contingency analysis, optimal power flow, short-circuit calculations, and topology searches—all grounded in engineering rather than pure text generation. To enhance performance, they introduced two novel algorithms: LLM-driven prompt refinement with human feedback and schema-adaptive hybrid retrieval-augmented generation (RAG) for accurate retrieval from large structured grid datasets. While the paper doesn’t use a traditional machine-learning dataset, evaluations were conducted using various power grid cases and user query scenarios to demonstrate that X-GridAgent can interpret diverse natural-language requests and generate interpretable, tool-invoked analytical results with high reliability and flexibility. Results show the system effectively automates interpretable power system analysis and handles previously unseen tasks in a modular, extensible way, significantly reducing manual effort and domain expertise requirements.

The authors in \cite{yang2025graphsearch} focused on improving GraphRAG (graph-based retrieval-augmented generation) by introducing GraphSearch, an agentic deep searching workflow that integrates dual-channel retrieval over both semantic text chunks and structural graph knowledge, with iterative, multi-turn reasoning and modular stages to better surface relevant evidence and leverage graph structure. GraphSearch enhances how large language models interact with and reason over structured graph data, addressing limitations of shallow retrieval and inefficient use of graph information. They evaluated GraphSearch across six multi-hop RAG benchmarks, demonstrating consistently higher answer accuracy and generation quality compared to traditional RAG strategies, confirming the effectiveness of their agentic approach to graph retrieval-augmented generation. A summary of recent graph, agent, and LLM integration frameworks is presented in Table~\ref{tab:agentic_models}.


\begin{table*}[t]
\centering
\caption{Simplified comparison of graph-, agent-, and LLM-based integration frameworks, highlighting key contributions and capabilities}
\small
\setlength{\tabcolsep}{5pt}
\renewcommand{\arraystretch}{1.2}
\begin{tabular}{p{3cm} p{2.5cm} p{5cm} p{5cm}}
\toprule
\textbf{Model} & \textbf{Category} & \textbf{Method Overview} & \textbf{Key Contribution} \\
\midrule

Agent-as-a-Judge \cite{you2026agent} & Evaluation & Multi-agent evaluation with planning, verification, and debate & Framework for reliable LLM evaluation \\

MAGMA \cite{jiang2026magma} & Memory & Multi-graph memory (semantic, temporal, causal) with guided traversal & Improves interpretability and long-context reasoning \\

Graph-S3 \cite{chang2025graph} & Graph Reasoning & Stepwise supervised retrieval using synthetic pipelines & Improves accuracy and F1 on QA benchmarks \\

AgenticMath \cite{liu2025agenticmath} & Data Generation & Multi-agent math data generation with filtering and CoT augmentation & Matches larger models via high-quality data \\

AriGraph \cite{anokhin2024arigraph} & Memory Graph & Dynamic KG with episodic + semantic memory updates & Outperforms RAG and full-history baselines \\

AgentBench \cite{xia2025experience} & Benchmark & Multi-task agent evaluation across planning and reasoning & Identifies long-horizon reasoning gaps \\

Self-Refine \cite{zhao2025agree} & Refinement & Iterative self-critique (Draft → Critique → Refine) & Improves generation quality \\

ToolStorm \cite{zhang2025llm} & Tools & Unified tool integration with API chaining & Enhances task success and composability \\

COT-PLUS \cite{dong2025s} & Reasoning & Structured planning with recursive decomposition & Outperforms standard CoT reasoning \\

GraphAgents \cite{stewart2026graphagents} & Scientific Agent & KG-guided multi-agent reasoning & Improves cross-domain problem solving \\

ReaGAN \cite{guo2025reagan} & Graph Learning & Node-level agent reasoning with GNN + RAG & Strong performance on graph benchmarks \\

GraphCodeAgent \cite{li2025graphcodeagent} & Code Generation & Dual-graph modeling for repository-level code generation & Outperforms repo-level RAG \\

AgentVNE \cite{zheng2026agentvne} & Graph RL & LLM + graph RL for agent placement & Reduces latency and improves acceptance \\

AgentEdge \cite{lu2026agentic} & Edge Systems & Multi-agent orchestration in edge-cloud environments & Improves efficiency and reduces API calls \\

\bottomrule
\end{tabular}
\label{tab:agentic_models}
\end{table*}

\section{Applications}

This section highlights real-world applications of graph-LLM integration across diverse domains. By combining the structured representation capabilities of graphs with the semantic reasoning power of large language models (LLMs), these systems enable more accurate, interpretable, and context-aware solutions to complex problems.

Graph-based representations provide explicit modeling of entities, relationships, and dependencies, while LLMs contribute flexible reasoning, natural language understanding, and generative capabilities. Their integration has proven particularly effective in domains where both structured knowledge and unstructured data must be jointly processed.

In this section, we review representative applications in key areas, including cybersecurity, healthcare, recommendation systems, and governance. For each domain, we discuss how graph-LLM approaches improve performance, enhance interpretability, and address domain-specific challenges. These examples demonstrate the practical impact of graph-LLM integration and highlight emerging trends in real-world deployments.

\subsection{Cybersecurity and Malware Analysis}
The integration of large language models (LLMs) with graph-based representations has emerged as a powerful paradigm in cybersecurity and malware analysis. Graph structures such as control-flow graphs, call graphs, and dependency graphs capture rich structural relationships that are difficult to obfuscate, while LLMs provide high-level semantic understanding and reasoning capabilities. By grounding language-based inference in explicit program and interaction structures, these hybrid approaches enhance robustness against code obfuscation, polymorphism, and rapidly evolving attack strategies.

Recent studies demonstrate that graph-enhanced LLM frameworks can uncover latent malicious behaviors by reasoning over structural patterns rather than relying solely on surface-level code features. For example, cluster-aware graph modeling combined with LLM-assisted recovery improves the detection of obfuscated JavaScript malware, while graph-guided vulnerability detection systems focus LLM reasoning on security-critical execution paths. These methods consistently improve detection accuracy and reduce false positives, while also offering interpretable insights into the underlying causes of security alerts.

Beyond malware detection, graph–LLM integration has been successfully applied to identifying adversarial and automated behaviors in online ecosystems. Graph-based representations of user interactions and network behaviors provide contextual signals that complement LLM-based semantic analysis, enabling the detection of coordinated or LLM-driven bot activity. Collectively, these advances highlight the effectiveness of graph-augmented LLMs as a scalable, explainable, and resilient foundation for next-generation cybersecurity solutions.

In \cite{liang2025breaking}, the authors focus on detecting malicious JavaScript code that has been intentionally obfuscated. They propose a hybrid method called DeCoda that combines large language model (LLM)-based deobfuscation with a cluster-aware code graph learning approach. The LLM is used to progressively reconstruct and normalize obfuscated JavaScript into Abstract Syntax Trees (ASTs), and then a hierarchical graph model with node-to-cluster attention captures both local and global structural relationships in the code graphs. The framework is evaluated on two benchmark malicious JavaScript datasets, where it achieves significantly higher detection performance (e.g., F1-scores around 94–97\%) compared with state-of-the-art baselines.\\

In \cite{lu2024grace}, the authors focus on improving software vulnerability detection by enhancing large language models (LLMs) with structural code information and in-context learning. They propose a framework called GRACE, which integrates graph structure representations (including code graph information such as abstract syntax trees, program dependence graphs, and control flow graphs) into LLM prompting and uses a demonstration selection module that retrieves similar code examples based on semantic, syntactic, and lexical similarity. The enhanced vulnerability detection module combines domain knowledge, graph-structure prompts, and in-context examples to guide the LLM in identifying vulnerabilities more accurately. The method is evaluated on three widely studied vulnerability detection benchmark datasets (Reveal, FFmpeg+Qemu, and Big-Vul), and GRACE significantly outperforms existing token-based and graph-based baselines in terms of F1 score.

In \cite{duan2025llm}, the authors focus on evaluating and enhancing the use of large language models (LLMs) for software vulnerability detection in source code. They systematically analyze how state-of-the-art LLMs can detect security flaws in C/C++ and other languages by comparing different prompting strategies, such as basic vulnerability queries, CWE-specific prompts, and data-flow-analysis-based prompts, to guide the models’ reasoning on code semantics. The evaluation is conducted on multiple popular code vulnerability datasets (including Juliet, OWASP Juliet, CVEFixes, and others) to measure detection performance and F1-scores across vulnerability types. The results highlight both the strengths and limitations of current LLM methods for vulnerability detection and suggest directions for improving prompt design and context-aware reasoning.

\subsection{Healthcare and Biomedical Knowledge Graphs}

In healthcare and biomedical domains, the integration of LLMs with knowledge graphs (KGs) addresses key challenges related to data heterogeneity, domain specificity, and interpretability. Biomedical knowledge is inherently relational, involving complex interactions among genes, diseases, drugs, phenotypes, and clinical outcomes. Knowledge graphs provide a structured representation of these relationships, allowing LLMs to perform reasoning that is grounded in curated and evidence-based medical knowledge.

Recent systems illustrate how KG-enhanced LLMs can support advanced clinical and research tasks. Web-scale biomedical knowledge graphs enable interactive exploration of treatment pathways, molecular mechanisms, and clinical evidence, supporting decision-making in precision medicine. Similarly, LLM-powered biomedical assistants leverage phenotypic and genomic graphs to facilitate structured querying, hypothesis generation, and knowledge discovery across large and diverse biomedical datasets.

Graph-empowered LLMs have also shown promise in mental health assessment and disease monitoring. By modeling symptoms, behavioral signals, and clinical indicators as interconnected graph elements, these approaches improve the reliability and contextual awareness of LLM-based predictions. Overall, the synergy between knowledge graphs and LLMs enhances transparency, factual consistency, and trustworthiness, which are essential for real-world healthcare and biomedical applications.

\begin{figure*}
    \centering
    \includegraphics[width=0.99\linewidth]{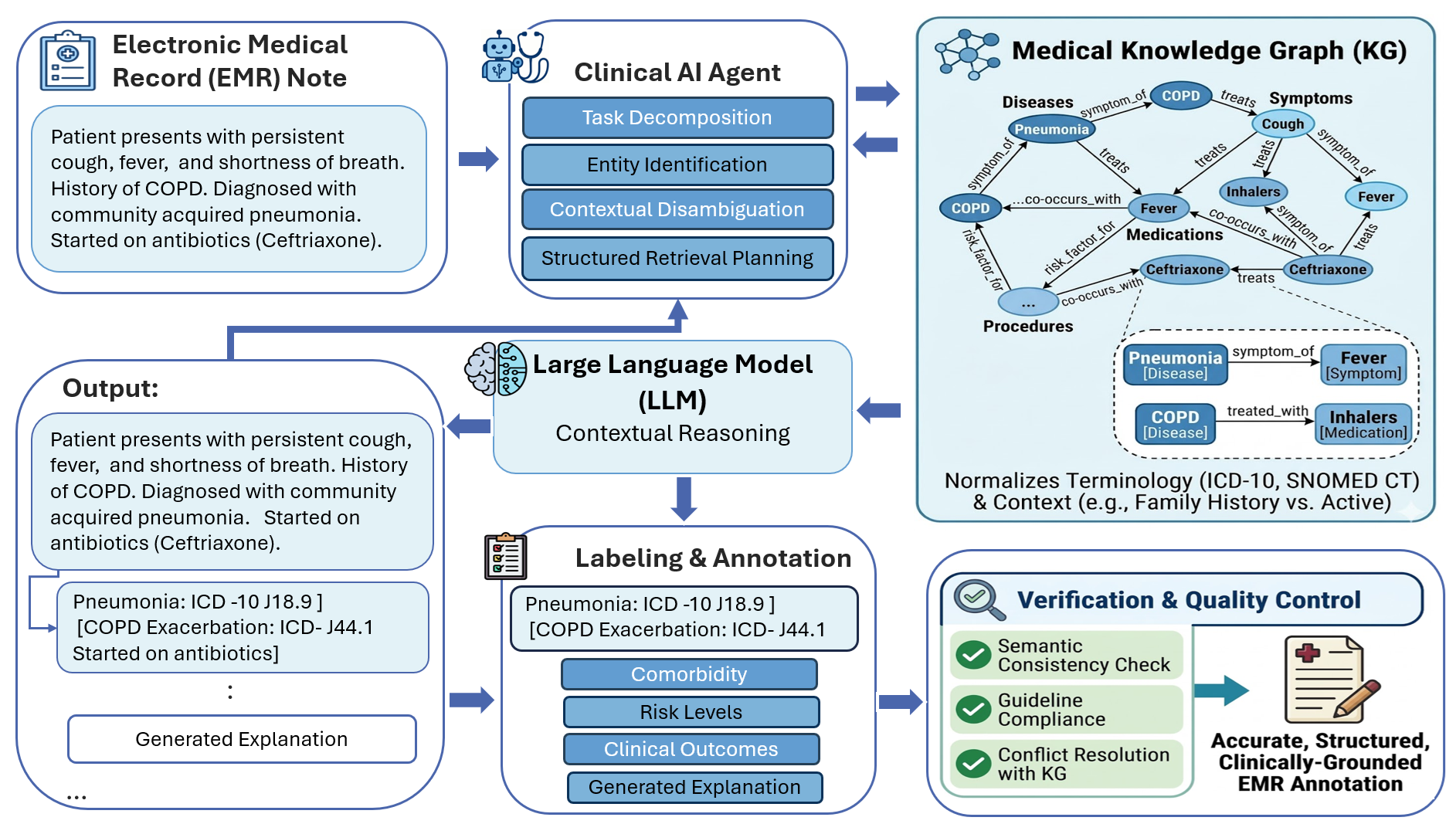}
    \caption{Architecture of Graph–Agent–LLM integration for Electronic Medical Record (EMR) labeling, including clinical agent reasoning, knowledge graph grounding, and verification modules}
    \label{fig:placeholder}
\end{figure*}

In \cite{gubanov2024cancerkg}, the authors focus on building a web-scale hybrid knowledge graph and large language model (KG-LLM) system to support interactive assistance with optimal cancer treatment and care. They propose CancerKG.ORG, an integrated platform that automatically ingests and organizes the latest peer-reviewed medical knowledge into a verifiable knowledge graph and combines it with an LLM for retrieval and reasoning. The knowledge graph acts as a guardrail to reduce hallucinations and ensure that responses are grounded in verified medical facts. The system is evaluated on real-world cancer treatment and clinical information tasks, demonstrating its ability to improve information retrieval quality compared with standalone LLM or KG methods

In \cite{o2024phenomics}, the authors focus on enabling non-expert users to interactively query and explore a large biomedical knowledge graph using natural language and large language models (LLMs). They introduce Phenomics Assistant, a prototype chat-based interface that translates user questions into calls to the Monarch Knowledge Graph (a comprehensive biomedical KG integrating gene, disease, and phenotype data), uses LLMs to interpret and refine queries, and returns summarized, factual responses grounded in the KG. The system is evaluated on benchmark gene-disease association and gene alias query tasks, showing that LLMs augmented with KG access produce significantly more accurate answers compared with standalone LLM responses

In \cite{chen2024depression}, the authors focus on automatic depression detection from clinical interview recordings by modeling the full multi-modal interview data (questions, answer transcripts, audio, and video) as a structural element graph (SEGA). They introduce SEGA, a directed acyclic graph representation aligned with human expertise about interview elements, and combine it with principle-guided LLM-based data augmentation and graph contrastive learning to address data scarcity. The method is evaluated on two real-world depression corpora (in English and Chinese), and results show that SEGA significantly outperforms strong baselines and powerful LLMs like GPT-3.5 and GPT-4 on depression classification tasks.

\subsection{Recommendation Systems}

Recommendation systems increasingly adopt graph-enhanced LLMs to overcome the limitations of traditional collaborative filtering and neural recommendation approaches. User–item interactions, social relationships, and contextual dependencies are naturally represented as graphs, capturing both direct and higher-order relational information. When combined with the reasoning and conversational capabilities of LLMs, these graph structures enable more personalized, adaptive, and context-aware recommendations.

Recent research demonstrates that graph-augmented LLMs effectively model long-range dependencies and evolving user intent, particularly in conversational recommendation scenarios. Knowledge graphs and interaction graphs guide LLM reasoning by providing structured representations of user preferences, item attributes, and historical behaviors. This integration improves performance in sparse-data settings and enhances robustness to noisy or incomplete interaction histories.

Moreover, graph–LLM integration significantly improves explainability in recommendation systems. By tracing paths and relationships within knowledge graphs, LLM-based recommenders can generate transparent and user-aligned explanations for their suggestions. These capabilities position graph-enhanced LLMs as a key enabler for trustworthy, interactive, and human-centered recommendation platforms.

In \cite{wei2024llmrec}, the authors focus on improving recommender systems under sparse feedback by leveraging large language models (LLMs) to augment user-item interaction graphs. They propose LLMRec, a framework that uses three simple LLM-based graph augmentation strategies reinforcing interaction edges, enhancing item attribute understanding, and generating user profile information, together with a denoising and robustification mechanism to refine augmented data. The approach is evaluated on benchmark recommendation datasets (e.g., MovieLens and Netflix), showing that LLM-augmented graphs lead to better recommendation performance compared with state-of-the-art baselines

In \cite{chen2023graph}, the authors focus on reviewing the capabilities and current limitations of graph neural networks (GNNs) when applied to knowledge graph representation and reasoning. They systematically analyze why standard GNN architectures struggle with tasks like long-range reasoning, multi-hop inference, and compositional generalization over symbolic relational structures. The paper discusses theoretical limitations (e.g., expressiveness bounds) and practical challenges (e.g., scalability and heterogeneity) and surveys existing extensions and alternatives aimed at bridging the gap between GNNs and symbolic relational models. This work does not introduce a new dataset or a new algorithm but synthesizes empirical and theoretical findings from prior studies to outline future research directions.

\subsection{Education, Law, and Governance}

In education, law, and governance, graph-enhanced LLMs address the need for structured reasoning over complex, rule-driven, and highly interconnected knowledge sources. Legal frameworks, educational curricula, and policy documents are naturally suited to graph representations, where entities and regulations are linked through formal relationships. Integrating these graphs with LLMs enables accurate question answering and decision support grounded in authoritative and structured knowledge.

Recent applications show that graph-augmented LLM systems can effectively support regulatory compliance and legal analysis. For instance, graph-enhanced question-answering systems for AI governance leverage structured representations of legal texts to guide LLM reasoning during regulatory transitions. Anchoring LLM outputs to graph-based legal knowledge reduces hallucinations and improves factual reliability in high-stakes decision-making environments.

In education and public administration, LLM-enabled knowledge graph construction facilitates scalable knowledge access and intelligent assistance. Cross-domain educational knowledge graphs allow LLMs to align student queries with structured learning objectives, while service-domain graphs support efficient governance-oriented information retrieval. These applications underscore the value of graph–LLM integration in promoting transparency, accountability, and informed decision-making across societal institutions.

In \cite{aggio2023graph}, the authors focus on creating a structured knowledge base that represents the legal concepts, obligations, risk categories, and regulatory requirements introduced by the European Union’s AI Act. They design an ontology-driven model to capture the semantics of the Act’s clauses, duties for different stakeholder roles (e.g., providers, deployers, users), and relationships between risk levels and compliance obligations. The method combines legal analysis, ontology engineering, and knowledge graph construction to formalize legal norms into a machine-readable representation. They build and publish the resulting AI Act Knowledge Base (AI-KB) and demonstrate its use with example queries that show how structured representations can support compliance checking, automated reasoning, and stakeholder guidance. The report does not use “datasets” in the machine-learning sense but uses the text of the European AI Act and related legal documents as its primary source to extract and encode concepts into a knowledge graph.

In \cite{bui2024cross}, the authors focus on building a knowledge graph from diverse educational data sources to support a large language model (LLM)-driven question-answering system in the context of higher education. They propose an automatic cross-data knowledge graph construction method that integrates structured and unstructured educational content (e.g., relational databases, text, APIs) into a unified KG and demonstrate how this KG can be used in conjunction with an LLM (such as ChatGPT) to answer educational queries. The approach is evaluated through a case study at Ho Chi Minh City University of Technology (HCMUT) on educational QA tasks, showing the effectiveness of the constructed knowledge graph for improving answer relevance and organization.

In this work, the authors focus on constructing a large-scale, open service domain knowledge graph to support service computing research. They propose BEAR, a service domain KG built using a comprehensive, manually designed service ontology and a zero-shot LLM-based knowledge extraction framework. The method automatically generates ontology-guided prompts and leverages large language models to extract high-quality entities, relations, and attributes from unannotated and heterogeneous data sources without supervised training. The resulting knowledge graph contains over 133k entities, 169k relations, and 424k factual attributes, and its feasibility and richness are demonstrated through empirical analysis rather than traditional ML benchmark datasets.

\begin{table*}[t]
\centering
\caption{When, why, where, and what to use for different graph-LLM integration paradigms.}
\label{tab:llm_graph_when_why}
\renewcommand{\arraystretch}{1.2}
\begin{tabularx}{\textwidth}{l X X X X}
\toprule
\textbf{Paradigm} & \textbf{What is it} & \textbf{When to Use} & \textbf{Why Use It} & \textbf{Where It Is Applied} \\
\midrule

\textbf{LLM-Assisted Knowledge Graph Construction} &
LLMs extract entities and relations from text to automatically build knowledge graphs. &
When large amounts of unstructured documents need to be converted into structured knowledge. &
Reduces manual annotation and simplifies traditional multi-step extraction pipelines. &
Scientific literature mining, enterprise knowledge bases, biomedical knowledge graphs. \\

\midrule

\textbf{Graph-Enhanced LLM Reasoning (GraphRAG)} &
LLMs reason over retrieved graph structures instead of flat text documents. &
When tasks require multi-hop reasoning across related facts. &
Graphs preserve relationships between facts and improve reasoning consistency. &
Question answering, fact verification, domain-specific search systems. \\

\midrule

\textbf{Hybrid GNN–LLM Models} &
Combines Graph Neural Networks for structural learning with LLMs for semantic reasoning. &
When both structural graph patterns and textual semantics are important. &
Leverages strengths of both models: GNNs capture topology while LLMs understand language. &
Molecular property prediction, recommender systems, knowledge graph learning. \\

\midrule

\textbf{KGQA with LLM} &
LLMs interpret questions and retrieve answers from knowledge graphs. &
When users need natural language access to structured knowledge. &
Provides interpretable reasoning over structured knowledge bases. &
Search engines, enterprise assistants, legal and financial knowledge systems. \\

\midrule

\textbf{Scene Graph + LLM} &
LLMs reason over scene graphs representing objects and spatial relationships. &
When tasks require visual scene understanding and spatial reasoning. &
Scene graphs provide structured representations of visual environments. &
Robotics, autonomous navigation, visual question answering, AR/VR systems. \\

\midrule

\textbf{Graph–Agent–LLM Integration} &
LLMs act as agents that plan tasks and interact with graph-based memory or workflows. &
When solving complex multi-step tasks requiring planning and tool use. &
Agents enable autonomous reasoning, coordination, and workflow execution. &
Scientific discovery systems, automated analytics, engineering decision systems. \\

\bottomrule
\end{tabularx}
\end{table*}

\section{Open Challenges and Future Directions}

Based on the surveyed literature and the integration patterns analyzed in this work, several open challenges emerge that limit the practical deployment and theoretical understanding of graph-LLM systems. These challenges are closely tied to the design choices discussed throughout this survey, including graph construction, retrieval, reasoning, and hybrid GNN-LLM architectures.

\subsection{Scalability}

Scalability is a recurring challenge across nearly all graph-LLM integration paradigms reviewed in this survey. LLM-assisted knowledge graph construction methods (Section~3) often rely on document chunking, iterative prompting, and multi-stage validation pipelines, which incur high computational costs when applied to large corpora or continuously evolving data sources. Similarly, GraphRAG systems (Section~4.1) require graph-aware retrieval, subgraph extraction, or path expansion, which can become prohibitively expensive as graph size and reasoning depth increase.

Hybrid GNN-LLM models (Section~5) further amplify scalability concerns by combining graph neural message passing with LLM inference, leading to increased memory usage and inference latency. These limitations are particularly evident in large-scale knowledge graphs, program graphs in cybersecurity, and dynamic scene graphs in multimodal environments. Future research should focus on hierarchical graph abstraction, approximate or adaptive subgraph retrieval, and incremental graph updates to enable scalable reasoning without sacrificing structural fidelity.

\subsection{Graph-LLM Alignment}

A central challenge highlighted throughout this survey is the alignment between graph structure and LLM reasoning behavior. While graphs encode explicit relational constraints, LLMs operate in a latent language space and may not consistently respect graph topology, edge semantics, or logical dependencies. This misalignment is evident in several surveyed settings, including LLM-assisted KG construction (Section~3), where hallucinated relations or schema drift may occur, and LLM-augmented KGQA (Section~6), where generative reasoning can deviate from graph-grounded evidence.

Existing approaches address alignment through prompt engineering, graph serialization, or post-hoc verification; however, these solutions remain fragile and task-specific. Ensuring that LLM reasoning faithfully follows graph constraints, particularly in multi-hop and incomplete-graph scenarios, remains an open problem. Future directions include structure-aware tokenization, graph-constrained decoding, tighter neuro-symbolic coupling, and joint objectives that explicitly penalize graph-inconsistent generations.

\subsection{Benchmark Gaps}

Although several benchmarks for graph-LLM systems have been introduced (Section~10), the current evaluation landscape remains fragmented. Many benchmarks focus on isolated tasks such as node classification, link prediction, or factual QA, while under-representing complex reasoning scenarios emphasized in this survey, including multi-hop GraphRAG reasoning, dynamic knowledge graph updates, causal inference, and cross-modal scene graph understanding.

Moreover, most evaluations emphasize task accuracy while overlooking critical dimensions such as hallucination rates, faithfulness to graph evidence, reasoning path correctness, and robustness to graph incompleteness or noise. This gap is particularly evident in application-driven domains such as cybersecurity and healthcare, where reliability and interpretability are as important as predictive performance. Developing unified, multi-dimensional benchmarks that reflect real-world graph-LLM workloads remains an important direction for future research.

\subsection{Trustworthy and Explainable Graph-LLM Models}

Trustworthiness and explainability emerge as cross-cutting concerns across all surveyed applications. While graphs provide a natural substrate for interpretable reasoning through paths, subgraphs, and relational chains, LLM-based reasoning may still produce opaque or ungrounded explanations if not explicitly constrained. This issue is particularly critical in high-stakes domains such as cybersecurity, healthcare, law, and governance (Section~9).

Several works reviewed in this survey leverage graphs as guardrails to mitigate hallucinations, yet ensuring that generated answers are both correct and verifiably grounded in graph evidence remains challenging. Future research should emphasize faithful explanation generation, explicit reasoning trace extraction, uncertainty-aware inference, and graph-grounded verification mechanisms. Strengthening the trustworthiness of graph-LLM systems will be essential for their safe deployment in real-world decision-making pipelines.


\section{Conclusion}

This survey systematically examined the emerging landscape of graph-LLM integration, addressing the fundamental questions of when, why, where, and how graphs and large language models should be combined. By organizing prior work according to graph modality, integration strategy, and application objective, we provided a unified view of methods spanning LLM-assisted graph construction, graph-enhanced reasoning, hybrid GNN-LLM models, KGQA, scene graph understanding, and domain-specific applications.

Our analysis shows that graph-LLM integration offers clear advantages for tasks requiring structured reasoning, multi-hop inference, and factual grounding, but also introduces new challenges related to scalability, alignment, evaluation, and trustworthiness. Importantly, the effectiveness of graph-LLM methods depends on careful design choices rather than generic integration.

We believe that future progress in this field will be driven by tighter neuro-symbolic alignment, scalable graph-aware reasoning mechanisms, realistic evaluation benchmarks, and a strong emphasis on explainability and reliability. This survey aims to serve as both a reference and a design guide for researchers and practitioners seeking to build robust, structure-aware, and trustworthy graph-LLM systems.

\bibliographystyle{ieeetr}
 \bibliography{references}

\newpage

 





\vfill

\end{document}